\documentclass{article}
\usepackage{spconf}
\usepackage{cite}
\usepackage{amsmath,amssymb,graphicx,subfigure,adjustbox,multirow,hhline}

\newcommand{\ykt}{\widetilde{\yv}_k}
\newcommand{\xkt}{\widetilde{\xv}_k}
\newcommand{\wkt}{\widetilde{\wv}_k}

\long\def\comment#1{}

\newcommand{\beq}{\begin{equation*}}
\newcommand{\eeq}{\end{equation*}}

\newfont{\bbb}{msbm10 scaled 700}

\newfont{\bb}{msbm10 scaled 1100}

\newcommand{\hv}{{\bf h}}

\newcommand{\wv}{{\bf w}}

\newcommand{\xv}{{\bf x}}
\newcommand{\yv}{{\bf y}}

\newcommand{\Am}{{\bf A}}

\newcommand{\Wm}{{\bf W}}

\newcommand{\Xm}{{\bf X}}
\newcommand{\Ym}{{\bf Y}}

\newcommand{\Kc}{{\cal K}}

\newcommand{\Nc}{{\cal N}}
\newcommand{\Oc}{{\cal O}}

\newcommand{\Sc}{{\cal S}}

\newcommand{\Vc}{{\cal V}}

\newcommand{\lambdav}{\hbox{\boldmath$\lambda$}}

\title{Image Denoising Via Collaborative Support-agnostic recovery}

\name{Muzammil Behzad$^{1}$ \quad Mudassir Masood$^{1}$ \quad Tarig Ballal$^{2}$ \quad Maha Shadaydeh$^{3}$ \quad Tareq Y. Al-Naffouri$^{1,2}$
}
\address{\normalsize{$^{1}$King Fahd University of Petroleum \& Minerals (KFUPM), Dhahran, Eastern Province, Saudi Arabia}\\
\normalsize{$^{2}$King Abdullah University of Science \& Technology (KAUST), Thuwal, Makkah Province, Saudi Arabia}\\
\normalsize{$^{3}$Institute for Computer Science and Control, Hungarian Academy of Sciences, Budapest, Hungary}
\\ \normalsize{\{g201402660, mudassir\}@kfupm.edu.sa, \{tarig.ahmed,tareq.alnaffouri\}@kaust.edu.sa}, shadaydeh.maha@sztaki.mta.hu}

\begin{document}
\maketitle
\begin{abstract}
	In this paper, we propose a novel image denoising algorithm using collaborative support-agnostic sparse reconstruction. An observed image is first divided into patches. Similarly structured patches are grouped together to be utilized for collaborative processing. In the proposed collaborative schemes, similar patches are assumed to share the same support taps. For sparse reconstruction, the likelihood of a tap being active in a patch is computed and refined through a collaboration process with other similar patches in the same group. This provides very good patch support estimation, hence enhancing the quality of image restoration. Performance comparisons with state-of-the-art algorithms, in terms of SSIM and PSNR, demonstrate the superiority of the proposed algorithm.
\end{abstract}


\section{Introduction}
\label{Introduction}
Image denoising algorithms aim at restoring image information from recorded version contaminated by noise. The noise is generally assumed to be signal independent zero mean additive white Gaussian. Over the past few decades, this problem has been extensively studied and the field has witnessed a continuous growth resulting in a number of highly effective image denoising algorithms.

The majority of these algorithms exploit a basic, yet effective, patch-based approach. These algorithms can be broadly classified into two categories of spatial and transform domain methods. The algorithms in the category of spatial-domain methods operate directly on the image pixels. A number of spatial-domain algorithms have been proposed showing performance improvement for a variety of denoising scenarios (see \cite{7298595, 1703579, doi:10.1137/040616024}). The NL-means algorithm \cite{1467423} is one of the most popular spatial-domain algorithms. This algorithm replaces each pixel by a weighted average of all other pixels. The weights are calculated based on the similarity in the neighborhood of each pixel with that of the reference pixel.

On the other hand, transform-domain algorithms perform denoising through thresholding of the weak coefficients in some transformed domain \cite{1014998,5211131, doi:10.1142/S0219691305000701,6226423,7350835, 7298646}. In particular, a sparse representation of the image over a given dictionary is utilized to identify the noise components. Typically, compressed sensing (CS) algorithms are applied to recover the sparse coefficients (see \cite{5653853, 6319405, ipol.2013.16, 7331303, 7351377}). One of the important denoising algorithms belonging to the transform-domain category is the K-SVD \cite{elad2006image}, which adapts a highly overcomplete dictionary computed via some prior training process. Unfortunately, this process impairs the computational flexibility of this algorithm.

Other denoising approaches combine more than one denoising methodology (see e.g., \cite{7025163,6820766,6678291, 7351639}), imposing additional computational burden. For example, BM3D \cite{4271520} takes advantage of both the spatial and transform-domain techniques by grouping similar image blocks in 3D arrays and applying collaborative filtering in the transform domain. In general, state-of-the-art algorithms are capable of denoising images with high precision. However, in some instances, these algorithms tend to produce over-smoothed images where important information like edges and textural details are lost.

In this paper, we propose a novel image denoising algorithm based on collaborative CS in a sparse transform domain. The proposed algorithm is named \emph{collaborative support-agnostic recovery} (CSAR). In the proposed algorithm, the sparse coefficients of an image patch are computed and refined via collaboration with similarly structured patches. This collaboration process in computing the supports of the patches results in a more accurate sparse representation of these patches, hence producing an enhanced image denoising performance. The proposed algorithm also lends itself well to computationally-simple implementation, as will be demonstrated in the following sections of this paper.

\section{The proposed collaborative support-agnostic recovery (CSAR)}
\label{ProposedAlgorithm}
Let $\Xm \in \mathbb{R}^{R\times C}$ be an image matrix. We aim to estimate the latent image matrix $\Xm$ from its noisy observations $\Ym$
\begin{align}\label{eq:11}
\Ym &= \Xm + \Wm,
\end{align}
where, $\Wm$ represents the noise matrix whose entries are i.i.d. random variables drawn from a Gaussian distribution with zero mean and variance $\sigma_w^2$. To find the estimated and denoised image $\widehat{\Xm}$, we use the following three main steps.

\subsection{Formation and grouping of image patches}\label{sec:grouping}
We form $N \times N$ square patches around each pixel in the image where $N$ is selected to be an odd number.\footnote{Our algorithm applies to the general case where patches could be rectangular or even linear. However, for simplicity and convenience we focus on the special case of square patches in this paper.} Further, to accommodate the border pixels, we pad the image borders with $\lfloor \frac{N}{2} \rfloor$ pixels. This results in a total number of $K=RC$ patches as
\begin{align}\label{eq:Ypatch}
\Ym_k = \Xm_k + \Wm_k, \quad \forall k \in \Kc.
\end{align}
where $\Kc = \{1,2,\dots, K\}$. Note that for computational convenience, we represent patches in (\ref{eq:Ypatch}) in vectorized form and use the resulting notation in the rest of the paper. So we have
\begin{align}\label{eq:Ypatch_vec}
\ykt = \xkt + \wkt, \quad \forall k \in \Kc.
\end{align}
where $\ykt, \xkt$ and $\wkt, \forall k$ are vectors of length $N^2$.

\begin{figure}[b!]
	\centering
	\includegraphics[width=\linewidth, height=7.85cm]{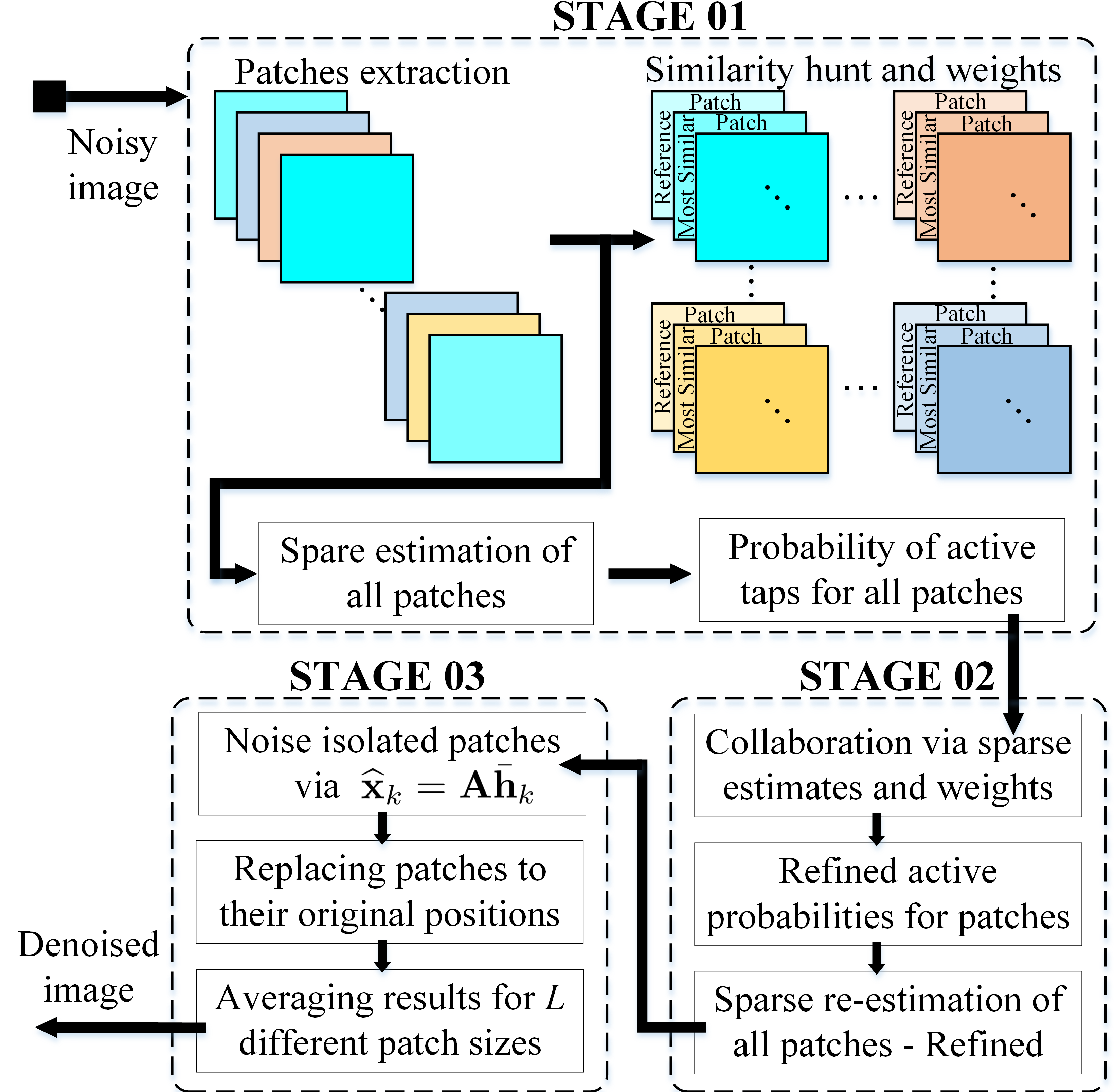}
	\caption{Flowchart of proposed CSAR denoising algorithm}
	\label{fig:fig31}
\end{figure}

The next step is to group each patch with similar patches as shown in stage 1 of Fig. \ref{fig:fig31}. The aim is to group all patches with similar underlying image structure irrespective of their intensity levels. Importantly, this intensity-invariant grouping requires normalization of the image patches as follows

\begin{align}
\yv_k = \eta(\ykt) =
\begin{cases}
\frac{\ykt}{\|\ykt\|}, & \quad \|\ykt\| \ne 0\\
\ykt, & \quad \text{otherwise}
\end{cases}, \quad \forall k \in \Kc,
\end{align}
where $\eta(\cdot)$ represents the normalization operator and $\yv_k$ is the normalized version of $\ykt$. As a result, we have
\begin{align}\label{eq:Ypatch_vec_normalized}
\yv_k = \xv_k + \wv_k, \quad \forall k \in \Kc.
\end{align}
Thus patch $\yv_k$ and those among all other patches that lie within a distance of, say $\epsilon$, from $\yv_k$ are grouped together. We call these the \emph{neighbors} of the $k$th patch i.e., $\yv_k$.  Thus
\begin{align}
\Nc_k = \{i:d(\yv_k, \yv_i) \le \epsilon\}, \quad \forall k \in \Kc
\end{align}
denotes a set of indices of all neighbors of patch $k$ and the index $k$ itself. Here $d(\cdot)$ could be any feasible distance measure, such as the Euclidean distance. Note that by virtue of the definition above, the set of neighbors need not be spatial neighbors and the set of neighbors $\Nc_k, \forall k\in \Kc$ are not disjoint. The upshot of such grouping is that it yields a higher number of neighbors for each patch which is beneficial for our collaborative approach as described in the following section.

\subsection{Collaborative Denoising}
It is a well-known fact that images are sparse in the wavelet domain. We use this property to find sparse representation of each patch as follows
\begin{align}
\yv_k & = \Am \hv_k + \wv_k, \quad \forall k \in \Kc
\end{align}
where $\Am \in \mathbb{R}^{N^2 \times M}, \, M \gg N^2$ is an overcomplete wavelet dictionary. Moreover, $\hv_k \in \mathbb{R}^{M}$  is the sparse representation of $\xv_k$ i.e., $\xv_k = \Am \hv_k$. Let $\widehat{\hv}_k$ represent an estimate of the sparse vector obtained through a sparse recovery algorithm and let  $\Sc_k$ be its support set. Note that in an ideal scenario $\Sc_k = \Sc_i, \forall i\in \Nc_k$ should hold true for all $k \in \Kc$. This observation motivates us to use the sparse representation of patches to devise a collaborative denoising method. However, note that in reality the supports may not match exactly as $\Nc_k$ is a function of a non-zero $\epsilon$ as well as $\wv_k$. The threshold $\epsilon$ could be selected such that it guarantees high similarity among the group members. However, the perturbations due to noise would remain and result in a disagreement among the supports of similar patches. Here we would like to stress that this disagreement is a blessing in disguise. Given sufficiently small $\epsilon$, most of the outliers $\Vc_k = \bigcup_{i\in\Nc_k} \Sc_i \backslash \bigcap_{i\in\Nc_k} \Sc_i$ in the support are there, with high probability, due to noise. This helps us identify and take care of the noise-causing components in the estimate $\widehat{\hv}_k$. One naive approach could be to eliminate the non-zero components of $\widehat{\hv}_k$ located at $\Vc_k$ and use the resulting sparse vector to form an estimate $\widehat{\xv}_k = \Am \widehat{\hv}_k$. However, this could result in destroying useful information especially in high noise cases as some legitimate non-zero locations could be mistaken for noise-causing components. In view of this, we resort to a much moderate approach.

In this approach, we utilize active probabilities of the non-zero locations of $\hv_k$. The idea is that similar patches will have similar support and the legitimate non-zero locations among these will have high active probabilities. Thus we propose that collaboration among patches take place in the sparse domain as shown in stage 2 of Fig. \ref{fig:fig31}. Specifically, for the $k$th patch, let $\lambdav_k\in\mathbb{R}^{M}$ represents the vector of active probabilities for the estimate $\widehat{\hv}_k$. We compute the weighted average\vspace{-0.1cm}
\begin{align}\label{eq:est_active_prob}
\lambdav_k^\prime &=
\frac{1}{\Nc_k}\sum_{j\in \Nc_k} \alpha_{j,k} \lambdav_j, \quad \forall k \in \Kc
\end{align}
as an estimate of the active probability vector of \emph{clean} $\hv_k$. The weighting factor \vspace{-0.1cm}
\begin{align}
\alpha_{j,k} \propto \frac{1}{d(\yv_j, \yv_k)}, \quad j\ne k.
\end{align}
This simple process allows us to gracefully downgrade the contribution of solitary active taps while preserving the values for locations that are common to most of the patches in $\Nc_k$. Moreover, by virtue of the law of large numbers, we expect that (\ref{eq:est_active_prob}) will result in a good estimate especially because $|\Nc_k|$ is large due to the intensity-invariant grouping approach. The derived clean $\lambdav_k^\prime$ is a valuable piece of information which approximates the \emph{a priori} information about the active locations of true or clean sparse representation of the $k$th patch $\xv_k$. This \emph{a priori} information could be provided to a sparse recovery algorithm, as shown in stage 2 of Fig. \ref{fig:fig31}, to find an estimate of true $\hv_k$ (let us call it $\bar{\hv}_k$) and thus an estimate of true (and denoised) $k$th patch which we denote as $\widehat{\xv}_k$
\begin{align}
\widehat{\xv}_k =
\begin{cases}
\eta^{-1}(\Am \bar{\hv}_k) = \Am \bar{\hv}_k \|\ykt\| \quad & \|\ykt\| \ne 0\\
\eta^{-1}(\Am \bar{\hv}_k) = \Am \bar{\hv}_k \quad & \text{otherwise}
\end{cases}, \quad \forall k \in \Kc.
\end{align}

\vspace{-0.7cm}
\subsection{Formation of final denoised image}

As described in Sec. \ref{sec:grouping}, we form overlapping patches. As a result each image pixel is present in $N^2$ patches and therefore has as many estimated values. In order to reconstruct the denoised image $\widehat{\Xm}$, we simply average the $N^2$ estimates of each pixel. In this way, the final image formation adds another level of averaging out impurities. Lastly, we average the denoising results using $L$ different odd patch sizes, stage 3 of Fig. \ref{fig:fig31}, that significantly improves the denoising performance.

\vspace{-0.1cm}
\section{Sparse recovery algorithm Selection} \vspace{-0.1cm}
\label{sparse_recovery}
Our denoising algorithm requires estimation of sparse vectors $\widehat{\hv}_k$ and $\bar{\hv}_k$. Several sparse recovery algorithms exist. However, we need to be careful in our selection. Specifically, the nature of our problem dictates that such algorithm should:
\begin{itemize}\vspace{-0.2cm}
	\item not pose strict conditions on the dictionary matrix $\Am$,\vspace{-0.3cm}
	\item be able to estimate parameters such as sparsity and variance of unknown vectors if not provided, \vspace{-0.3cm}
	\item be invariant to the distribution of unknown, and \vspace{-0.3cm}
	\item be capable of utilizing any available \emph{a priori} info.
\end{itemize}\vspace{-0.2cm}
Many algorithms exist that offer these features. However, very few have all of the mentioned attributes. Among these algorithms, we are specially interested in SABMP \cite{6581876} as it is capable of MMSE estimation even if the distribution of the unknown vector is not available. Moreover, the algorithm provides active probabilities along with the estimated sparse vector which is what our denoising algorithm benefits from.

\section{Computational complexity of our Image denoising algorithm}
\label{complexity}
\begin{figure}[b]
	\centering
	\subfigure{\includegraphics[width=4.1cm, height=4cm]{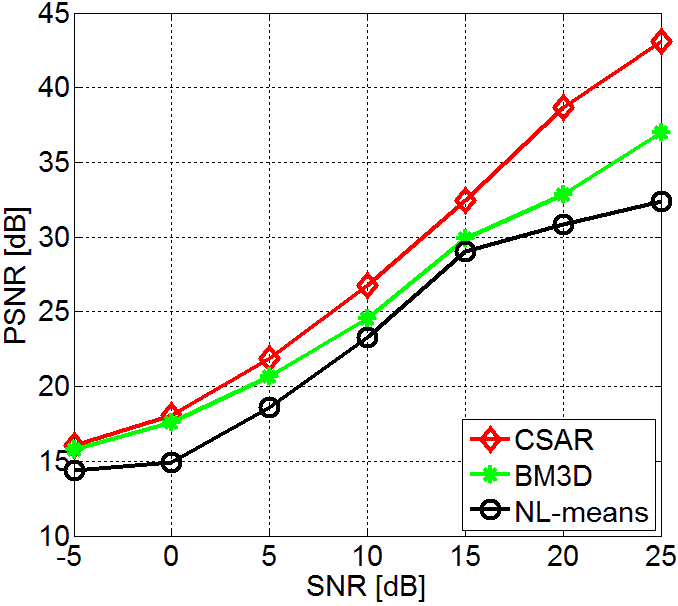}}\quad
	\subfigure{\includegraphics[width=4.1cm, height=4cm]{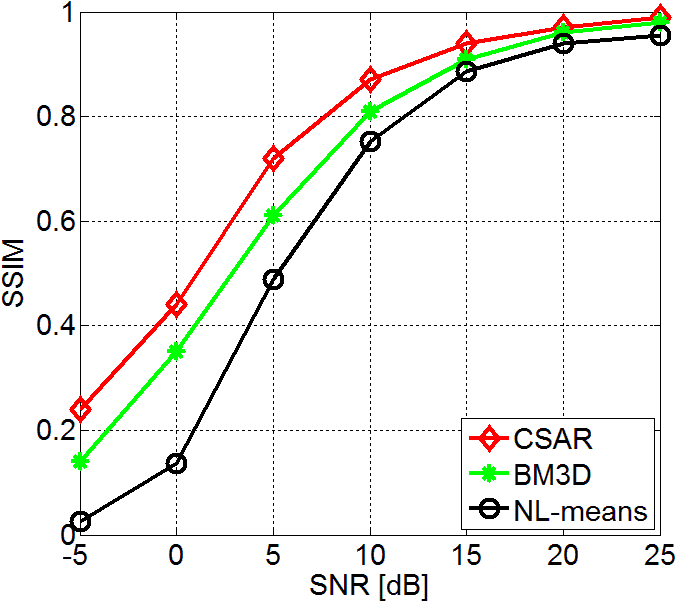}}
	\caption{PSNR and SSIM comparison of \textit{Peppers} image}
	\label{fig:fig2}
\end{figure}
The computational complexity of the proposed denoising approach is dominated by the computational complexity of the sparse recovery algorithm, which fortunately has a low computational complexity as compared to many similar algorithms. Given the dimensions of our problem, the computational complexity of estimating one $\hv_k$ through SABMP is of order $\Oc(MN^2P)$ where $P$ is the expected number of non-zeros (usually a very small number). Finally, for all $K$ patches and $L$ iterations for different patch sizes, the complexity will scale to an order of $\Oc(KLMN^2P)$.
\begin{figure*}[t]
	\centering
	\subfigure{\includegraphics[width=2.8cm, height=2.48cm]{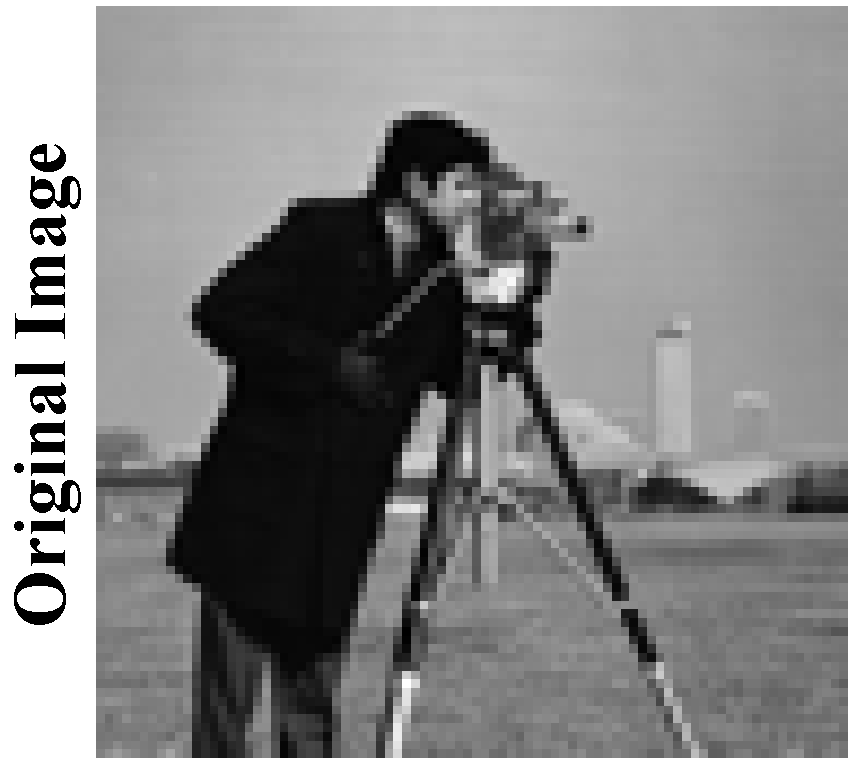}}\quad
	\subfigure{\includegraphics[width=2.8cm, height=2.8cm]{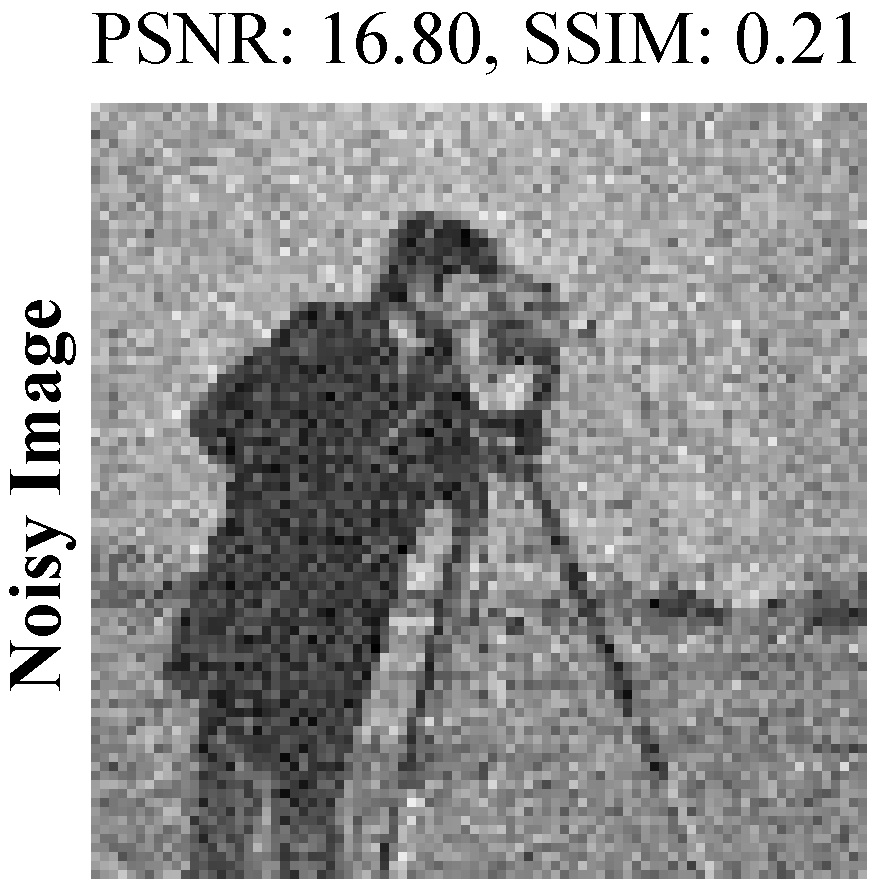}}\quad
	\subfigure{\includegraphics[width=2.8cm, height=2.8cm]{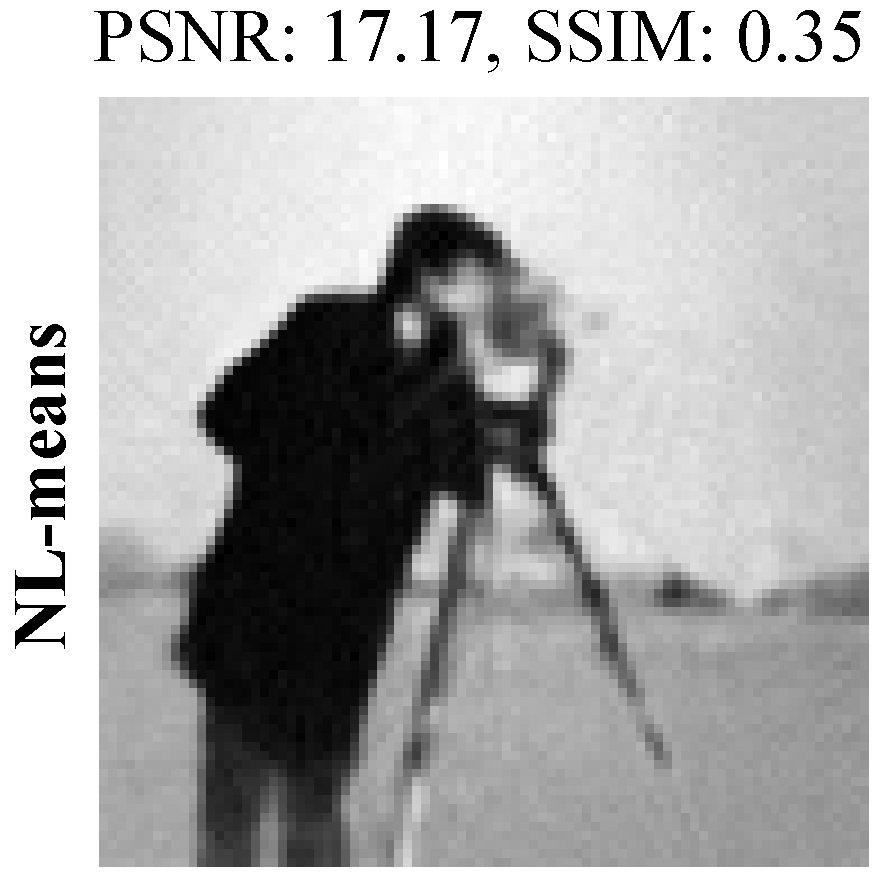}}\quad
	\subfigure{\includegraphics[width=2.8cm, height=2.8cm]{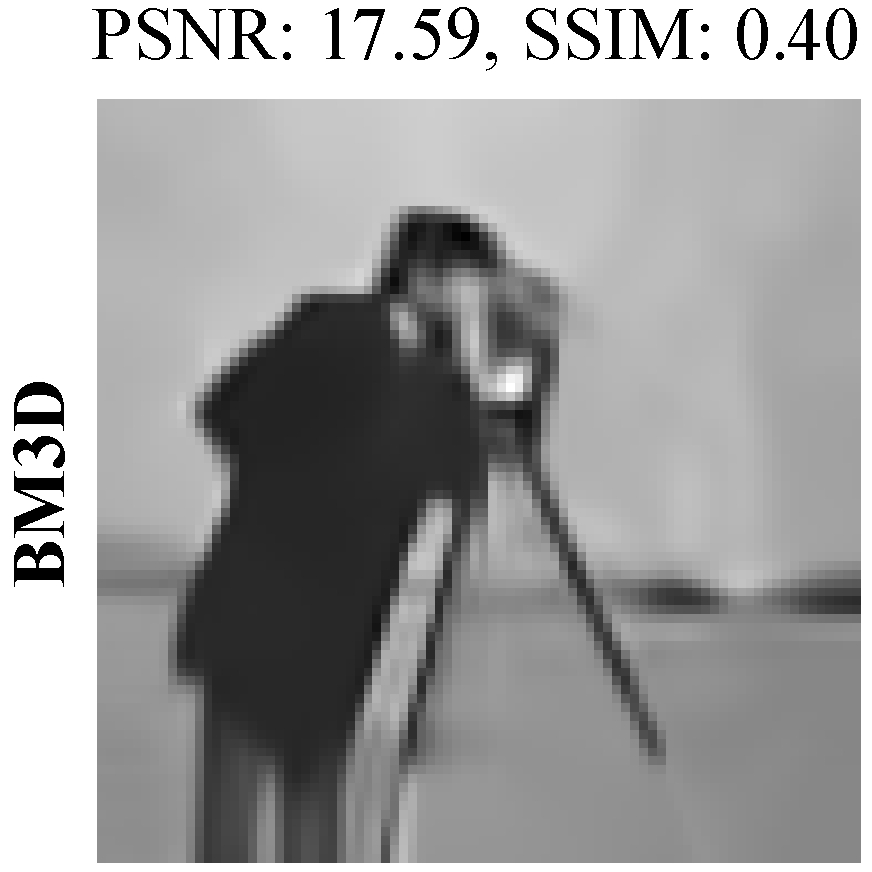}}\quad
	\subfigure{\includegraphics[width=2.8cm, height=2.8cm]{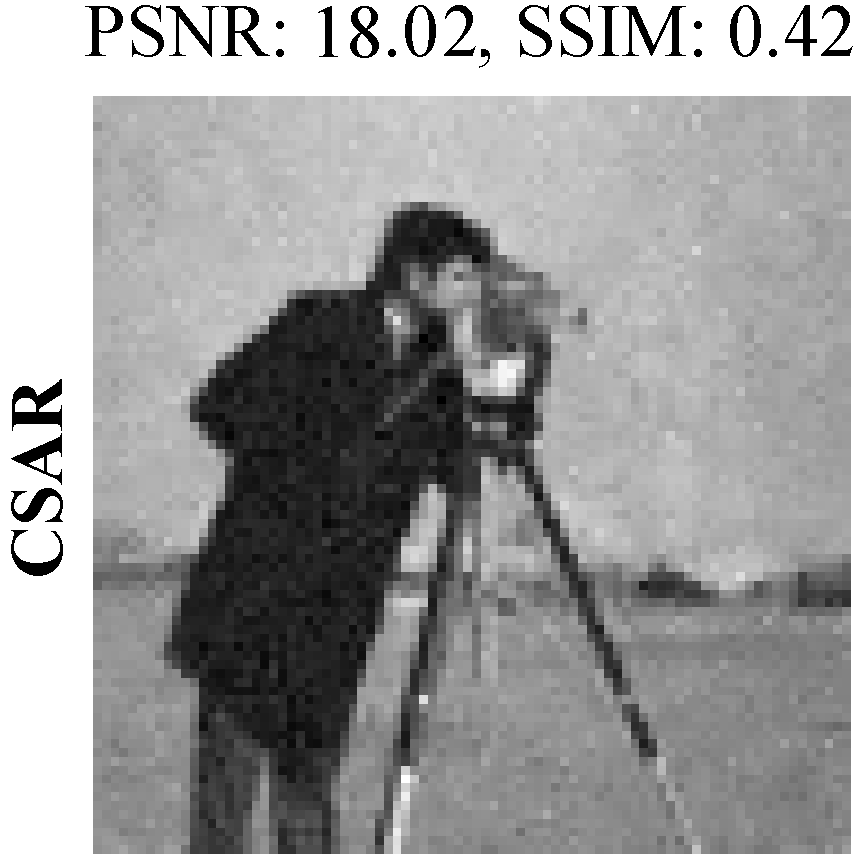}}
	\caption{Left to right: original and noisy \textit{Cameraman}, denoised by: NL-means, BM3D and CSAR at SNR$_{dB}$/$\sigma$ = 5/33}
	\label{fig:fig4}
\end{figure*}
\begin{figure*}[th!]
	\centering
	\subfigure{\includegraphics[width=2.8cm, height=2.48cm]{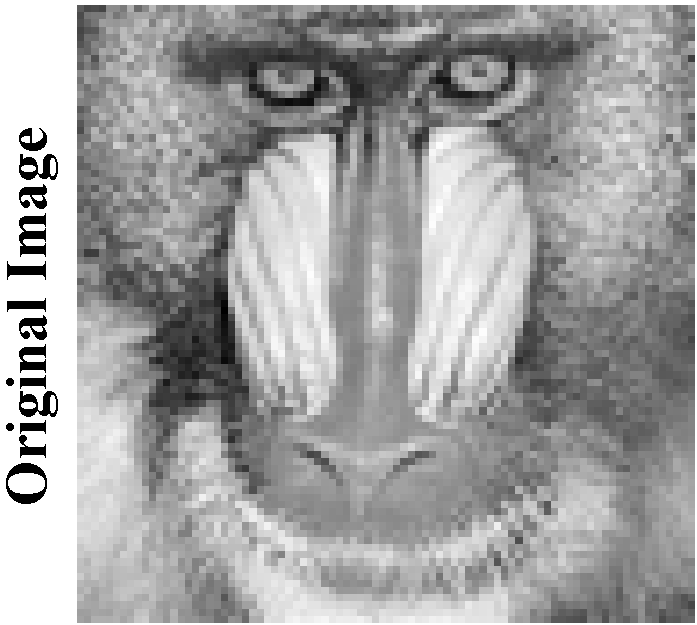}}\quad
	\subfigure{\includegraphics[width=2.8cm, height=2.8cm]{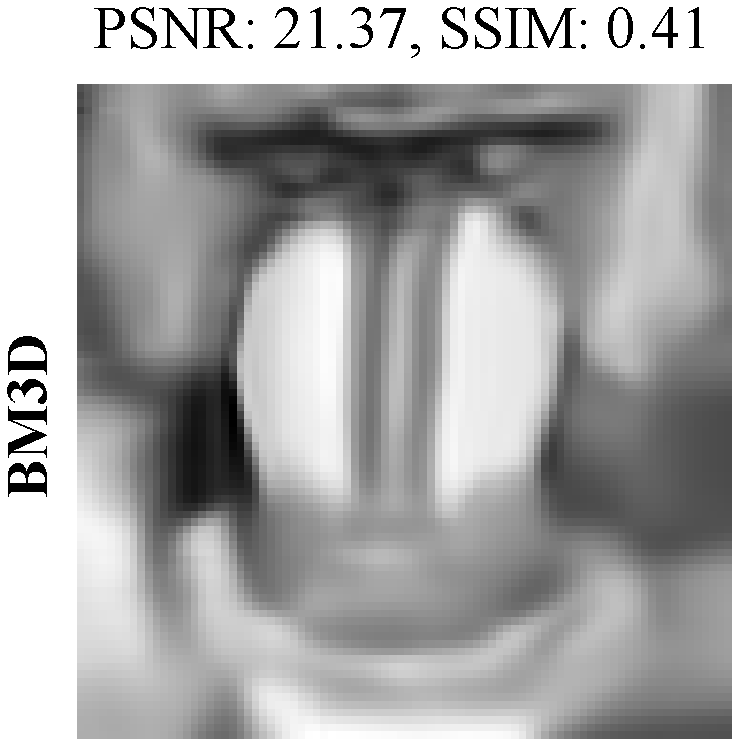}}\quad
	\subfigure{\includegraphics[width=2.8cm, height=2.8cm]{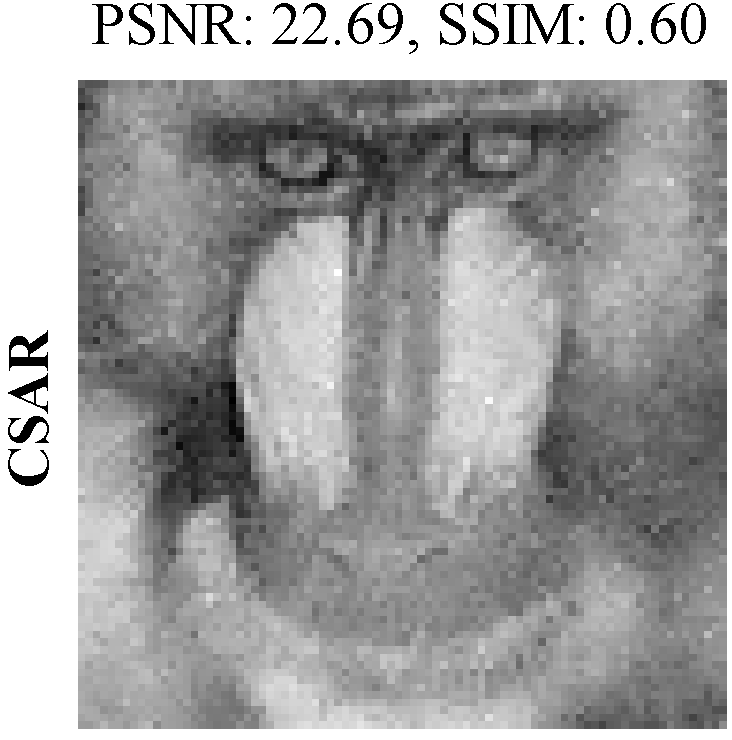}}\quad
	\subfigure{\includegraphics[width=2.8cm, height=2.8cm]{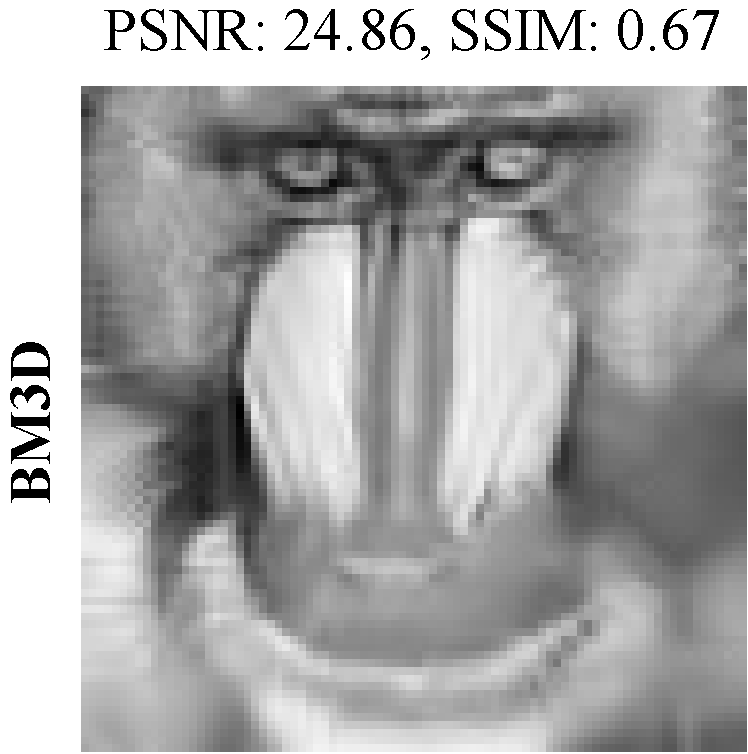}}\quad
	\subfigure{\includegraphics[width=2.8cm, height=2.8cm]{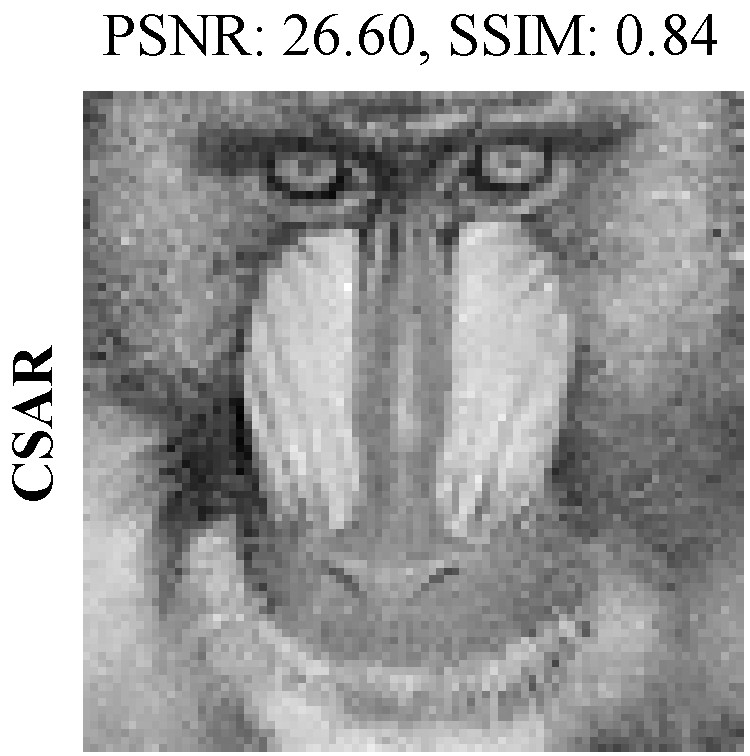}}
	\caption{Left to right: original \textit{Mandrill}, denoised by BM3D and CSAR at SNR$_{dB}$/$\sigma$ = 0/58 and 5/33}
	\label{fig:fig3}
\end{figure*}
\begin{table*}[t!]
	\centering
	\begin{tabular}{|c|c|c|c|c|c|c|c|c|c|c|}\hline\hline
		\multicolumn{2}{|c|}{SNR [dB] / $\sigma$} & Cameraman & Lena & Barbara & House & Peppers & Living Room & Boat  \\ \cline{1-9}
		\hline\hline
		\multirow{2}{*}{-5/103}
		& CSAR & \textbf{14.88/0.12} & \textbf{16.61/0.26} & \textbf{15.20/0.22} & \textbf{16.79/0.17} & \textbf{16.07/0.24} & \textbf{17.70/0.25} & \textbf{17.34/0.22}  \\ \cline{2-9}
		& BM3D & 14.73/0.10 & 16.37/0.19 & 14.87/0.11 & 16.07/0.11 & 15.80/0.14 & 16.64/0.14 & 16.95/0.09 \\ \hline
		
		\multirow{2}{*}{0/58}
		& CSAR & \textbf{16.57/0.24} & \textbf{19.32/0.50} & \textbf{16.99/0.43} & \textbf{19.29/0.32} & \textbf{18.06/0.44} & \textbf{19.81/0.45} & \textbf{19.49/0.41} \\ \cline{2-9}
		& BM3D & 16.57/0.22 & 18.94/0.43 & 16.86/0.34 & 19.26/0.28 & 17.60/0.35 & 18.60/0.29 & 17.97/0.24 \\ \hline
		
		\multirow{2}{*}{5/33}
		& CSAR & \textbf{18.02/0.42} & \textbf{24.95/0.77} & \textbf{21.13/0.75} & \textbf{24.39/0.50} & \textbf{21.86/0.72} & \textbf{23.24/0.74} & \textbf{21.97/0.67}  \\ \cline{2-9}
		& BM3D & 17.59/0.40 & 23.91/0.71 & 20.55/0.67 & 22.98/0.48 & 20.68/0.61 & 21.21/0.58 & 19.68/0.50  \\ \hline
		
		\multirow{2}{*}{10/18}
		& CSAR & \textbf{22.60/0.58} & \textbf{26.90/0.87} & \textbf{27.34/0.92} & \textbf{30.34/0.60} & \textbf{26.78/0.87} & \textbf{32.46/0.91} & \textbf{24.06/0.86}  \\ \cline{2-9}
		& BM3D & 22.32/0.57 & 25.04/0.86 & 26.40/0.88 & 28.37/0.60 & 24.53/0.81 & 29.24/0.86 & 22.47/0.76  \\ \hline
		
		\multirow{2}{*}{15/10}
		& CSAR & \textbf{27.45/0.71} & \textbf{28.97/0.91} & \textbf{35.22/0.97} & \textbf{37.78/0.70} & \textbf{32.47/0.94} & \textbf{36.92/0.96} & \textbf{25.53/0.94}  \\ \cline{2-9}
		& BM3D & 27.07/0.71 & 27.45/0.90 & 31.84/0.95 & 36.12/0.67 & 29.93/0.91 & 32.86/0.93 & 24.50/0.89 \\ \hline
		
		\multirow{2}{*}{20/6}
		& CSAR & \textbf{33.55/0.83} & \textbf{33.03/0.94} & \textbf{39.84/0.98} & \textbf{42.25/0.78} & \textbf{38.68/0.97} & \textbf{41.55/0.98} & \textbf{26.80/0.97} \\ \cline{2-9}
		& BM3D & 31.84/0.79 & 32.33/0.92 & 35.49/0.97 & 39.34/0.73 & 32.87/0.96 & 36.79/0.97 & 25.45/0.94 \\ \hline
		
		\multirow{2}{*}{25/3}
		& CSAR & \textbf{39.78/0.91} & \textbf{33.91/0.96} & \textbf{44.55/0.99} & \textbf{46.70/0.85} & \textbf{43.09/0.99} & \textbf{46.27/0.99} & \textbf{28.12/0.98} \\ \cline{2-9}
		& BM3D & 37.08/0.87 & 33.01/0.94 & 39.49/0.98 & 42.82/0.79 & 36.97/0.98 & 41.01/0.98 & 27.30/0.97 \\ \hline\hline
	\end{tabular}
	\caption{Comparison of denoising grayscale images using CSAR and BM3D both in terms of PSNR [dB] and SSIM}
	\label{table1}
\end{table*}

\section{Simulation Results and Discussions}
\label{Results}
In this section, we compare the proposed algorithm with two state-of-the-art algorithms, namely, NL-means \cite{1467423} and BM3D \cite{4271520}. Comparisons with NL-means and BM3D validate the superior performance of CSAR and prove that our algorithm is even robust to situations where these cannot perform well.

For the experiments, we used various grayscale standard test images. For a more challenging competition, an SNR range including very high noise levels were used providing higher chances of confusing signal components with noise. The entries of dictionary were derived from wavelet as well as DCT basis. Square patch sizes of 3, 5, 7 and 9, i.e., $L=4$, were used and the denoising results were averaged.

Fig. \ref{fig:fig2} compares the performance of denoising the peppers image by proposed CSAR with BM3D and NL-means algorithms. The peppers image is specifically selected for its \textit{detailed rich} nature making the comparison more interesting. It is obvious that the proposed algorithm outperforms the other two algorithms across the considered SNR range. Apart from outperforming in terms of PSNR, the SSIM performance of CSAR is also much better than other competing algorithms.

The comparison of denoising \textit{Cameraman} is provided in Fig. \ref{fig:fig4}. This experimental results taken at SNR = 5 dB depict that our algorithm outperforms state-of-the-art algorithms. Another comparison of \textit{Mandrill} at SNR = 0 dB and 5 dB is illustrated in Fig. \ref{fig:fig3}. These figures emphasize the importance of preserving feature rich portions, as done by CSAR, which are more likely to get destroyed in the presence of noise.\vspace{-0.05cm}

Specifically, we show in Fig. \ref{fig:fig4} that our results are not blurred at high noise of SNR = 5 dB, while in Fig. \ref{fig:fig3} we show that we are good at preserving the details. For instance in Fig. \ref{fig:fig3}, note that the face details are blurred out both at SNR = 0 and 5 dB in BM3D but exist in CSAR denoised image. This degradation due to blurring or removal of feature rich components can have critical consequences e.g. detecting tumors in bio-medical applications, that can be life threatening if detections go wrong. Detailed results are provided in Table \ref{table1} for a number of test images widely used in the denoising literature. These extensive results demonstrate the superiority and efficacy of our approach over images of different types.

\begin{figure*}[t!]
	\subfigure{\includegraphics[width=5.3cm, height=5.3cm]{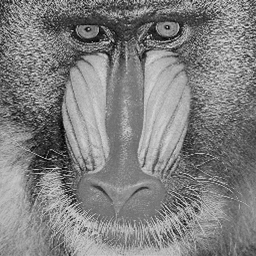}}\quad
	\subfigure{\includegraphics[width=5.3cm, height=5.3cm]{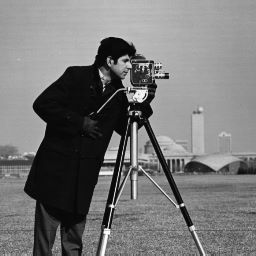}}\quad
	\subfigure{\includegraphics[width=5.3cm, height=5.3cm]{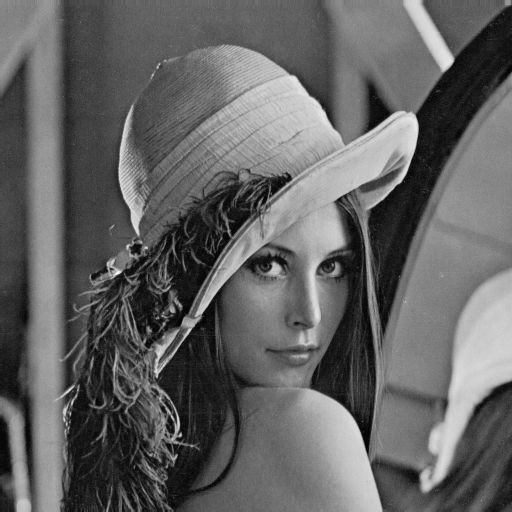}}\\
	\subfigure{\includegraphics[width=5.3cm, height=5.3cm]{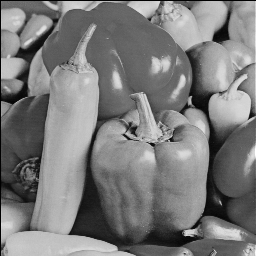}}\quad
	\subfigure{\includegraphics[width=5.3cm, height=5.3cm]{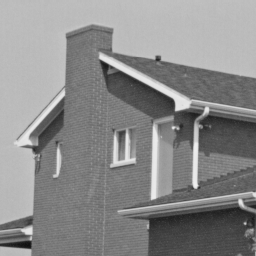}}\quad
	\subfigure{\includegraphics[width=5.3cm, height=5.3cm]{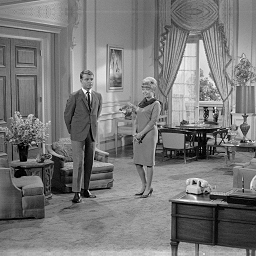}}\\	
	\subfigure{\includegraphics[width=5.3cm, height=5.3cm]{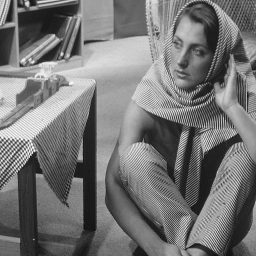}}\quad
	\subfigure{\includegraphics[width=5.3cm, height=5.3cm]{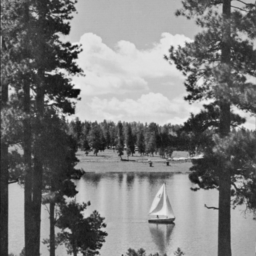}}\quad
	\subfigure{\includegraphics[width=5.3cm, height=5.3cm]{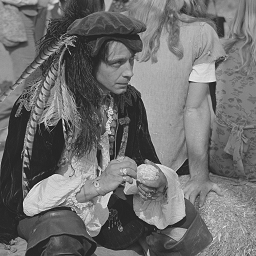}}\\	
	\subfigure{\includegraphics[width=5.3cm, height=5.3cm]{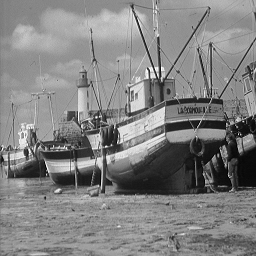}}
	\caption{First column top to bottom: original Mandrill, Peppers, Barbara and Boat images. Second column top to bottom: original Cameraman, House and Lake images. Third column top to bottom: original Lena, Living Room and Man images.}
	\label{all_original}
\end{figure*}

\begin{figure*}[t!]	
	\subfigure{\centering\includegraphics[width=0.49\linewidth]{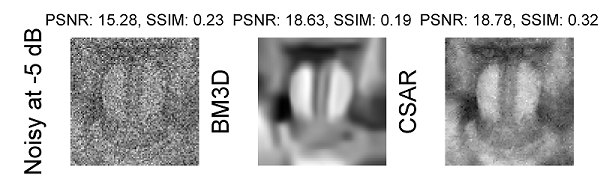}}\quad
	\subfigure{\centering\includegraphics[width=0.49\linewidth]{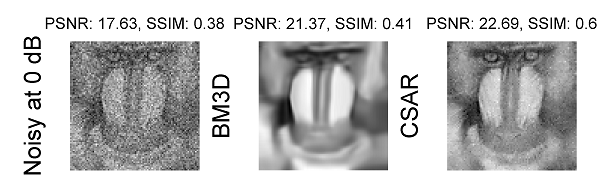}}\\
	\subfigure{\centering\includegraphics[width=0.49\linewidth]{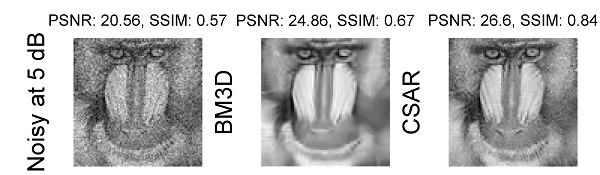}}\quad
	\subfigure{\centering\includegraphics[width=0.49\linewidth]{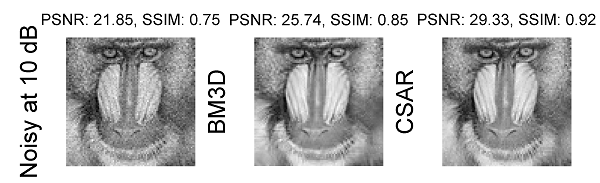}}\\
	\subfigure{\centering\includegraphics[width=0.49\linewidth]{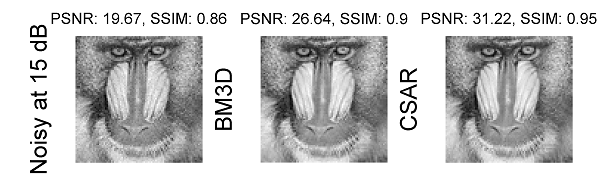}}\quad
	\subfigure{\centering\includegraphics[width=0.49\linewidth]{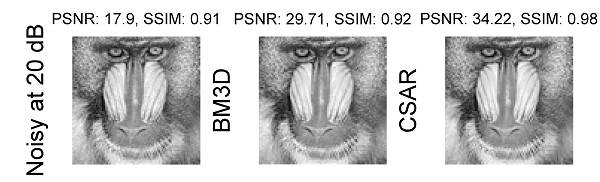}}\\
	\subfigure{\centering\includegraphics[width=0.49\linewidth]{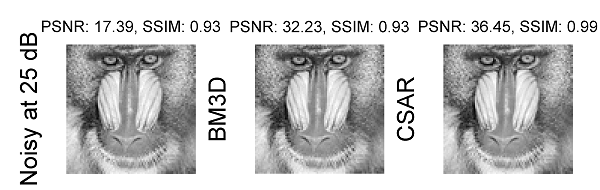}}\\
	\subfigure{\centering\includegraphics[width=0.49\linewidth]{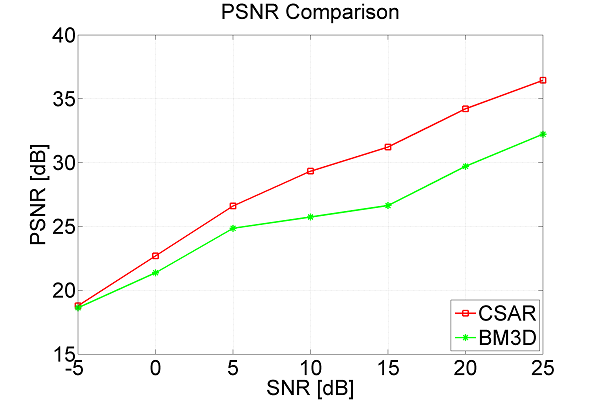}}\quad
	\subfigure{\centering\includegraphics[width=0.49\linewidth]{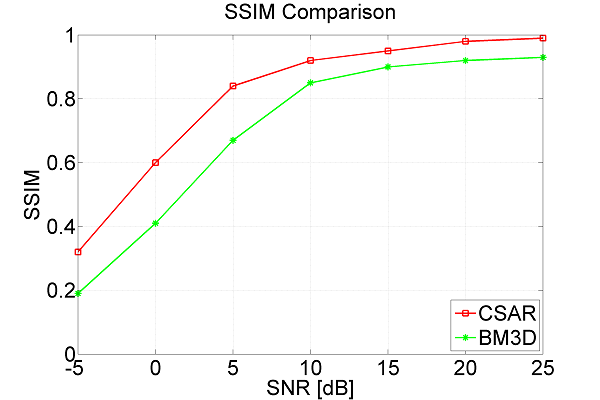}}
	\caption{A comparison of BM3D and CSAR algorithm's based denoising results of Mandrill standard test image over an extensive SNR range of -5 dB to 25 dB.}
	\label{R_mandrill}
\end{figure*}

\begin{figure*}[t!]	
	\subfigure{\centering\includegraphics[width=0.49\linewidth]{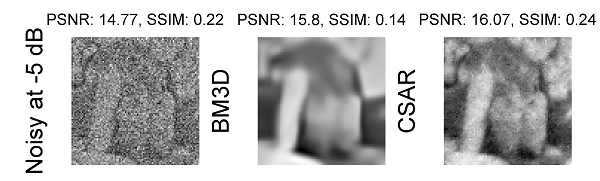}}\quad
	\subfigure{\centering\includegraphics[width=0.49\linewidth]{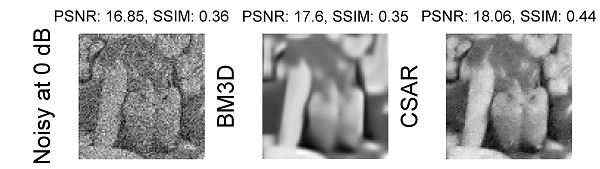}}\\
	\subfigure{\centering\includegraphics[width=0.49\linewidth]{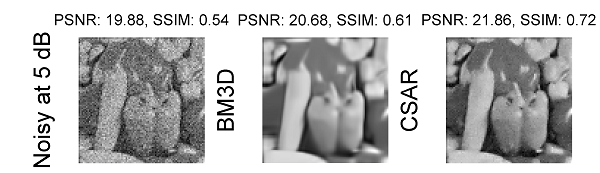}}\quad
	\subfigure{\centering\includegraphics[width=0.49\linewidth]{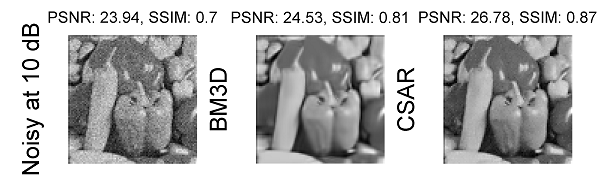}}\\
	\subfigure{\centering\includegraphics[width=0.49\linewidth]{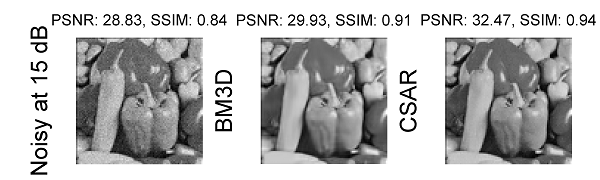}}\quad
	\subfigure{\centering\includegraphics[width=0.49\linewidth]{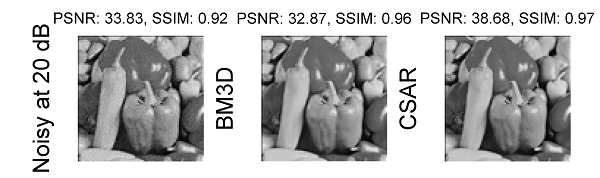}}\\
	\subfigure{\centering\includegraphics[width=0.49\linewidth]{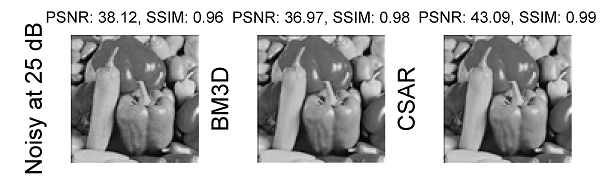}}\\
	\subfigure{\centering\includegraphics[width=0.49\linewidth]{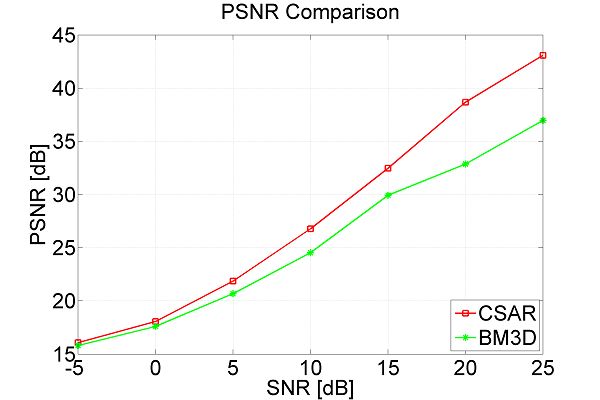}}\quad
	\subfigure{\centering\includegraphics[width=0.49\linewidth]{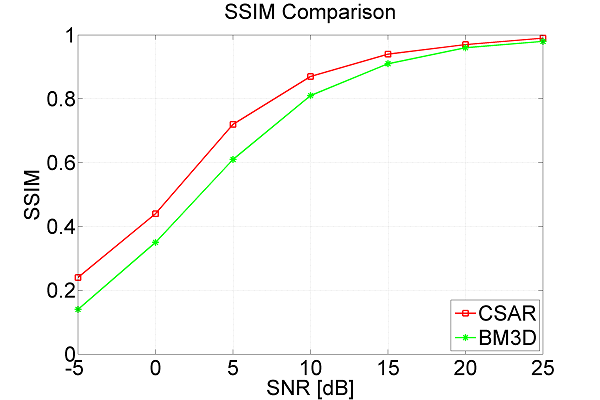}}
	\caption{A comparison of BM3D and CSAR algorithm's based denoising results of Peppers standard test image over an extensive SNR range of -5 dB to 25 dB.}
	\label{R_peppers}
\end{figure*}

\begin{figure*}[t!]	
	\subfigure{\centering\includegraphics[width=0.49\linewidth]{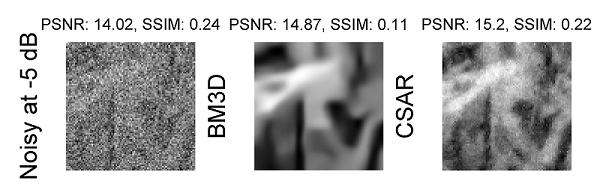}}\quad
	\subfigure{\centering\includegraphics[width=0.49\linewidth]{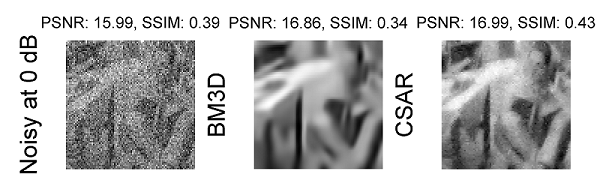}}\\
	\subfigure{\centering\includegraphics[width=0.49\linewidth]{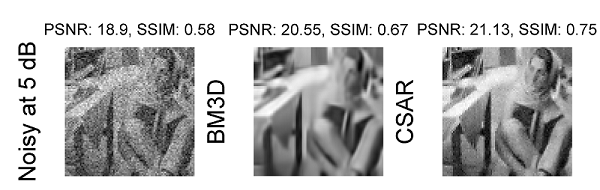}}\quad
	\subfigure{\centering\includegraphics[width=0.49\linewidth]{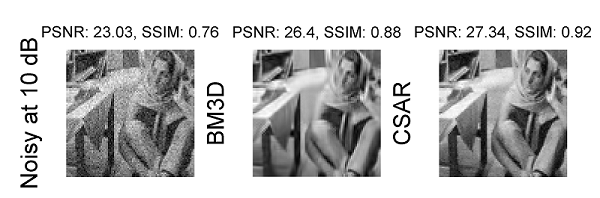}}\\
	\subfigure{\centering\includegraphics[width=0.49\linewidth]{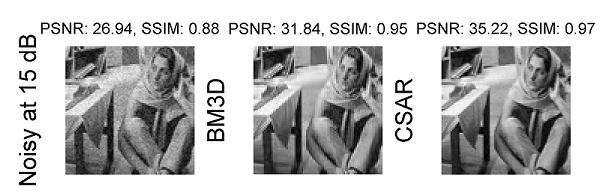}}\quad
	\subfigure{\centering\includegraphics[width=0.49\linewidth]{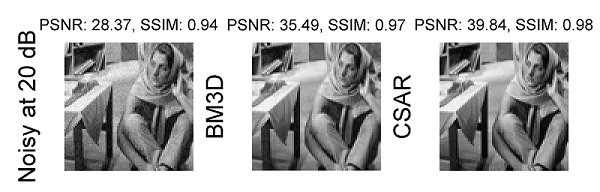}}\\
	\subfigure{\centering\includegraphics[width=0.49\linewidth]{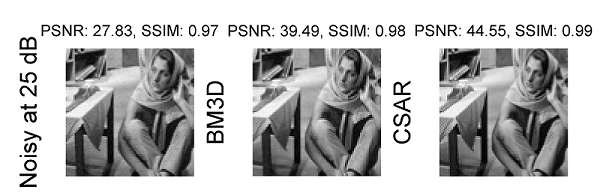}}\\
	\subfigure{\centering\includegraphics[width=0.49\linewidth]{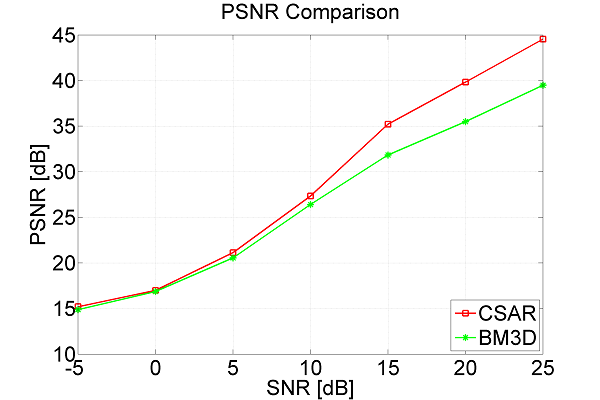}}\quad
	\subfigure{\centering\includegraphics[width=0.49\linewidth]{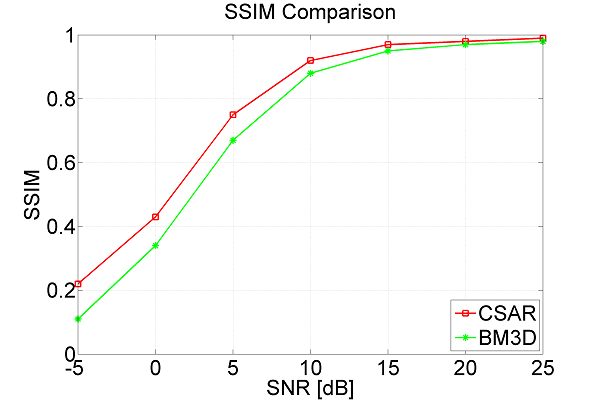}}
	\caption{A comparison of BM3D and CSAR algorithm's based denoising results of Barbara standard test image over an extensive SNR range of -5 dB to 25 dB.}
	\label{R_Barbara}
\end{figure*}

\begin{figure*}[t!]	
	\subfigure{\centering\includegraphics[width=0.49\linewidth]{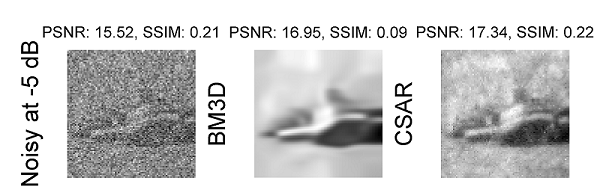}}\quad
	\subfigure{\centering\includegraphics[width=0.49\linewidth]{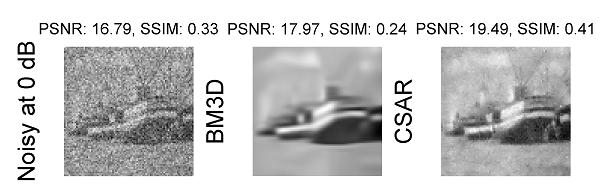}}\\
	\subfigure{\centering\includegraphics[width=0.49\linewidth]{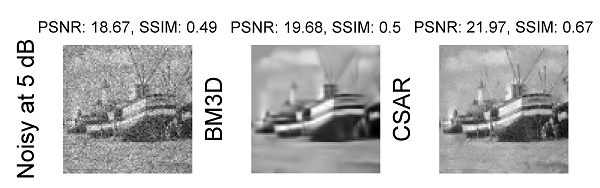}}\quad
	\subfigure{\centering\includegraphics[width=0.49\linewidth]{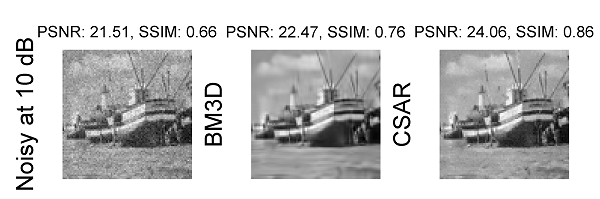}}\\
	\subfigure{\centering\includegraphics[width=0.49\linewidth]{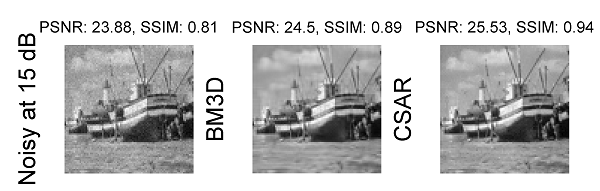}}\quad
	\subfigure{\centering\includegraphics[width=0.49\linewidth]{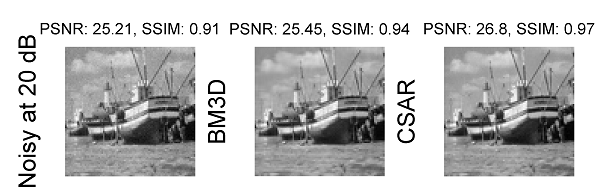}}\\
	\subfigure{\centering\includegraphics[width=0.49\linewidth]{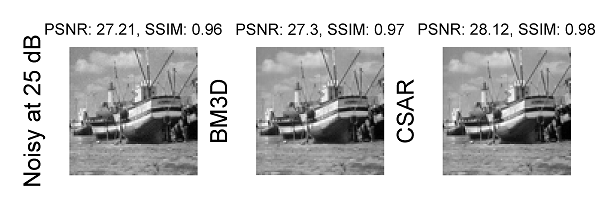}}\\
	\subfigure{\centering\includegraphics[width=0.49\linewidth]{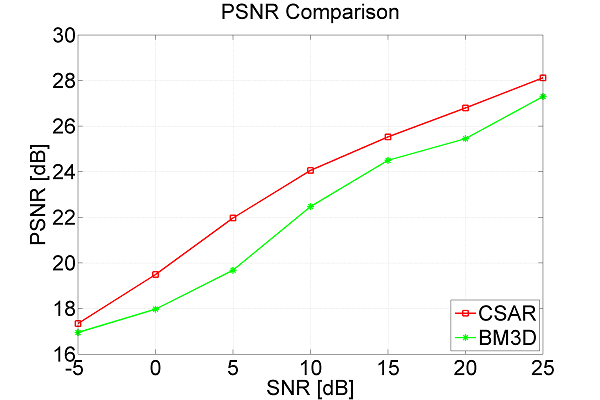}}\quad
	\subfigure{\centering\includegraphics[width=0.49\linewidth]{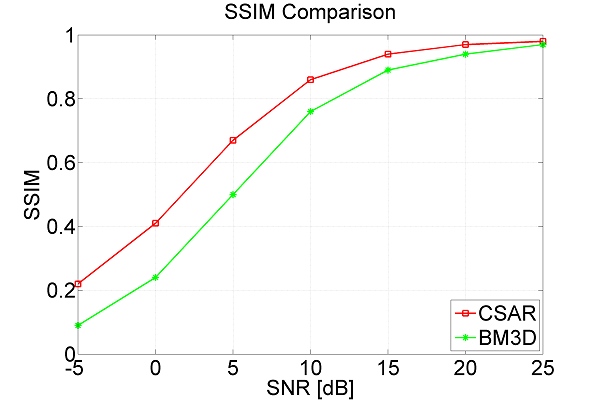}}
	\caption{A comparison of BM3D and CSAR algorithm's based denoising results of Boat standard test image over an extensive SNR range of -5 dB to 25 dB.}
	\label{R_Boat}
\end{figure*}

\begin{figure*}[t!]	
	\subfigure{\centering\includegraphics[width=0.49\linewidth]{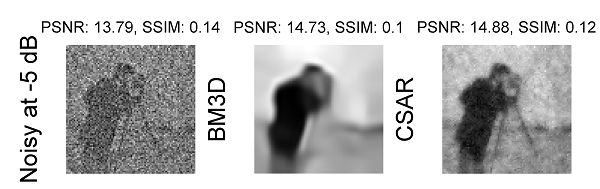}}\quad
	\subfigure{\centering\includegraphics[width=0.49\linewidth]{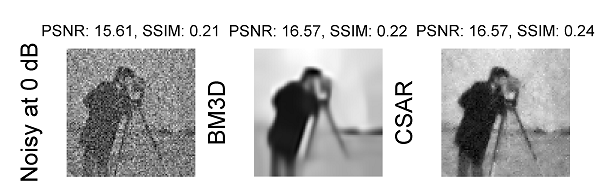}}\\
	\subfigure{\centering\includegraphics[width=0.49\linewidth]{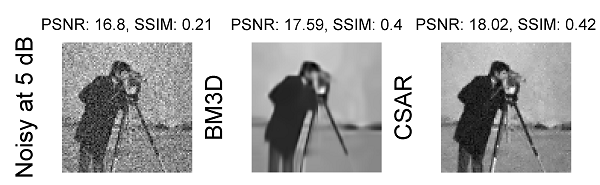}}\quad
	\subfigure{\centering\includegraphics[width=0.49\linewidth]{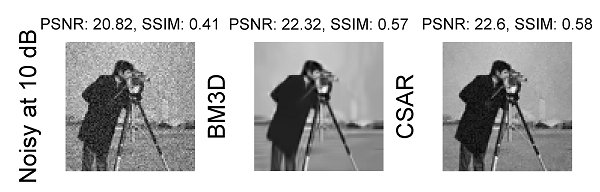}}\\
	\subfigure{\centering\includegraphics[width=0.49\linewidth]{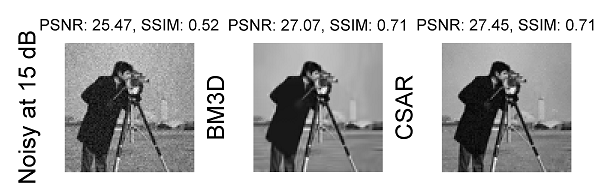}}\quad
	\subfigure{\centering\includegraphics[width=0.49\linewidth]{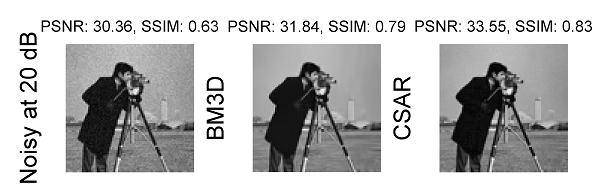}}\\
	\subfigure{\centering\includegraphics[width=0.49\linewidth]{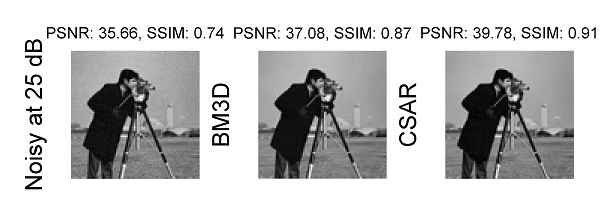}}\\
	\subfigure{\centering\includegraphics[width=0.49\linewidth]{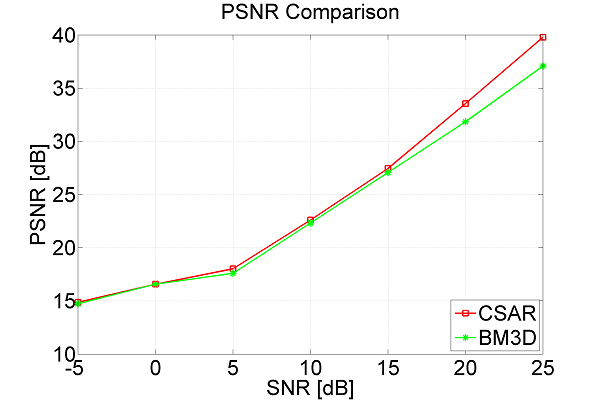}}\quad
	\subfigure{\centering\includegraphics[width=0.49\linewidth]{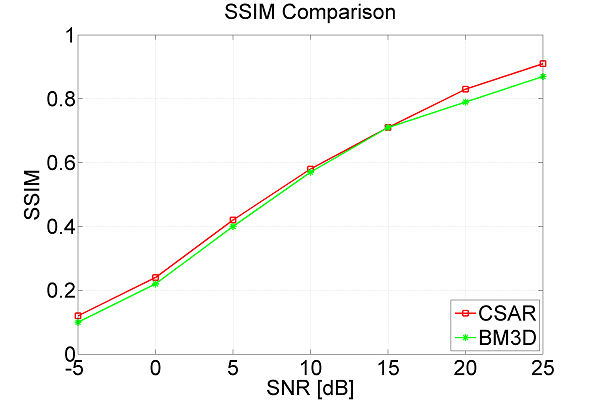}}
	\caption{A comparison of BM3D and CSAR algorithm's based denoising results of Cameraman standard test image over an extensive SNR range of -5 dB to 25 dB.}
	\label{R_Cman}
\end{figure*}

\begin{figure*}[t!]	
	\subfigure{\centering\includegraphics[width=0.49\linewidth]{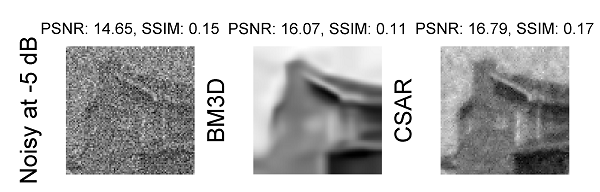}}\quad
	\subfigure{\centering\includegraphics[width=0.49\linewidth]{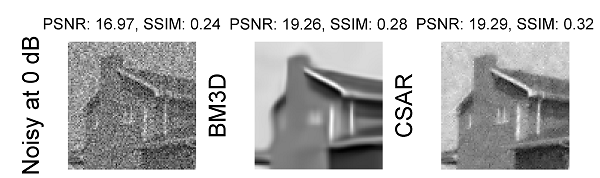}}\\
	\subfigure{\centering\includegraphics[width=0.49\linewidth]{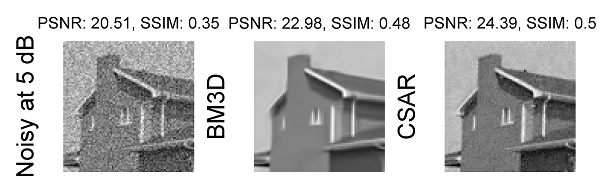}}\quad
	\subfigure{\centering\includegraphics[width=0.49\linewidth]{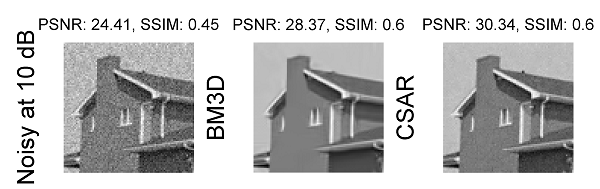}}\\
	\subfigure{\centering\includegraphics[width=0.49\linewidth]{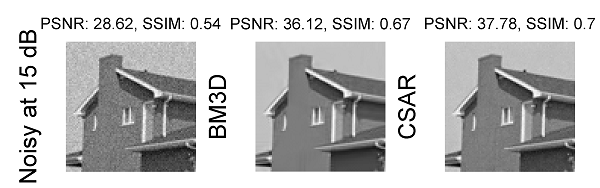}}\quad
	\subfigure{\centering\includegraphics[width=0.49\linewidth]{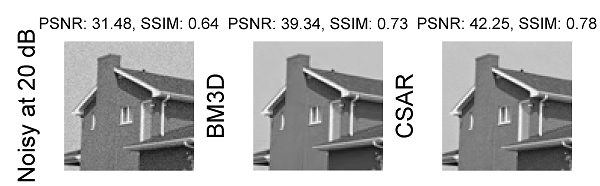}}\\
	\subfigure{\centering\includegraphics[width=0.49\linewidth]{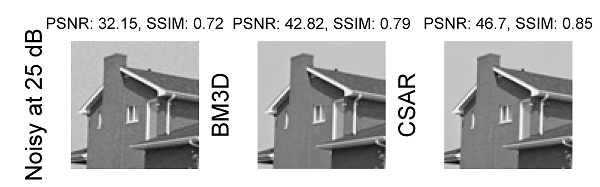}}\\
	\subfigure{\centering\includegraphics[width=0.49\linewidth]{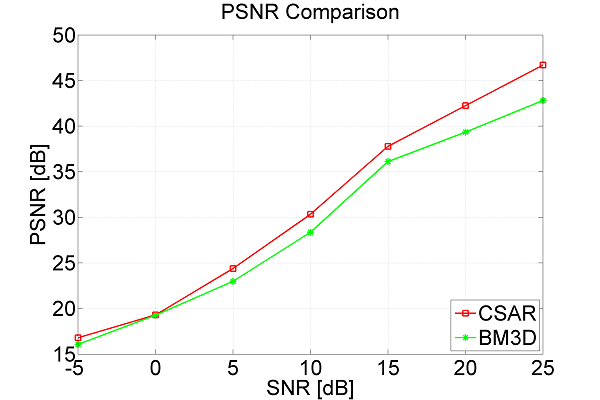}}\quad
	\subfigure{\centering\includegraphics[width=0.49\linewidth]{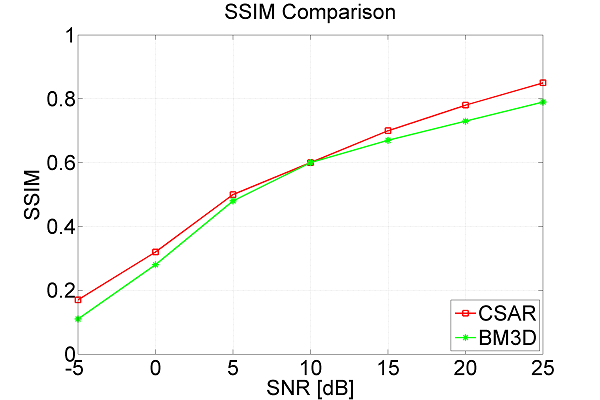}}
	\caption{A comparison of BM3D and CSAR algorithm's based denoising results of House standard test image over an extensive SNR range of -5 dB to 25 dB.}
	\label{R_House}
\end{figure*}

\begin{figure*}[t!]	
	\subfigure{\centering\includegraphics[width=0.49\linewidth]{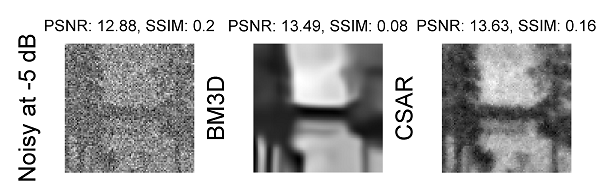}}\quad
	\subfigure{\centering\includegraphics[width=0.49\linewidth]{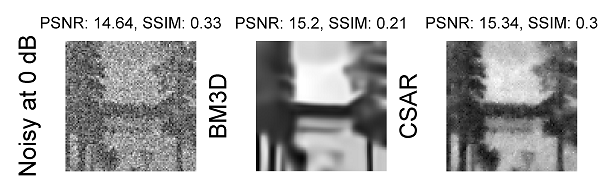}}\\
	\subfigure{\centering\includegraphics[width=0.49\linewidth]{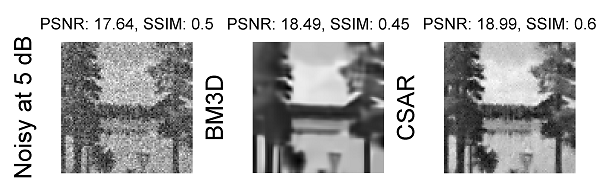}}\quad
	\subfigure{\centering\includegraphics[width=0.49\linewidth]{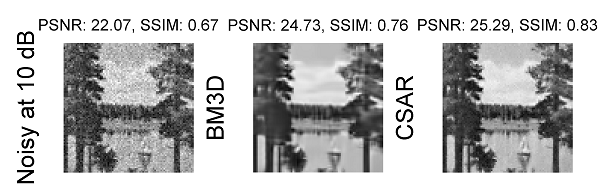}}\\
	\subfigure{\centering\includegraphics[width=0.49\linewidth]{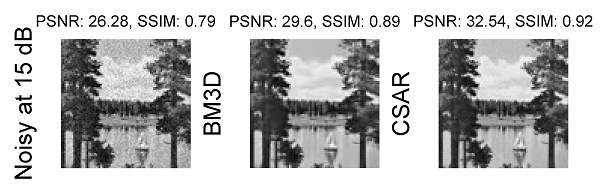}}\quad
	\subfigure{\centering\includegraphics[width=0.49\linewidth]{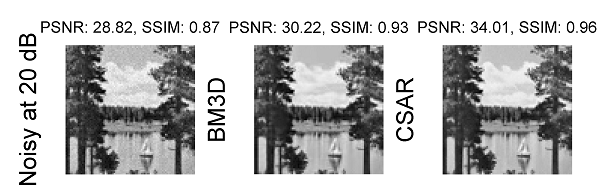}}\\
	\subfigure{\centering\includegraphics[width=0.49\linewidth]{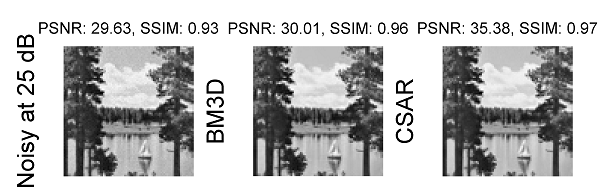}}\\
	\subfigure{\centering\includegraphics[width=0.49\linewidth]{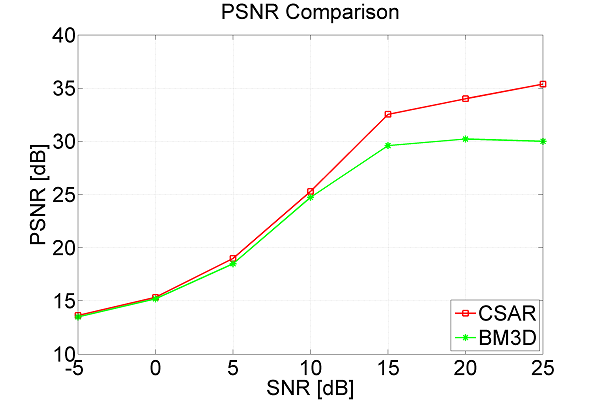}}\quad
	\subfigure{\centering\includegraphics[width=0.49\linewidth]{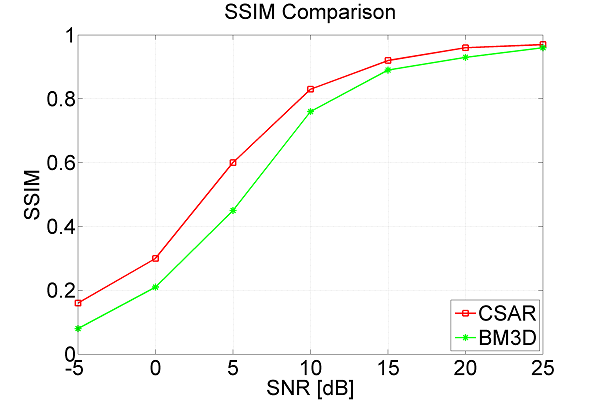}}
	\caption{A comparison of BM3D and CSAR algorithm's based denoising results of Lake standard test image over an extensive SNR range of -5 dB to 25 dB.}
	\label{R_Lake}
\end{figure*}

\begin{figure*}[t!]	
	\subfigure{\centering\includegraphics[width=0.49\linewidth]{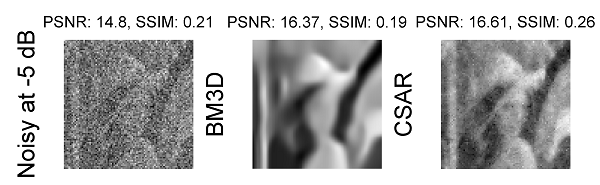}}\quad
	\subfigure{\centering\includegraphics[width=0.49\linewidth]{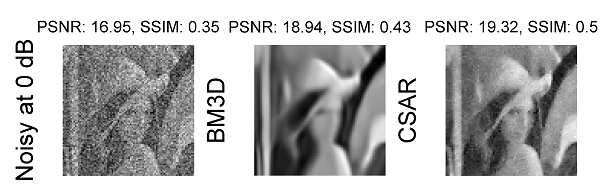}}\\
	\subfigure{\centering\includegraphics[width=0.49\linewidth]{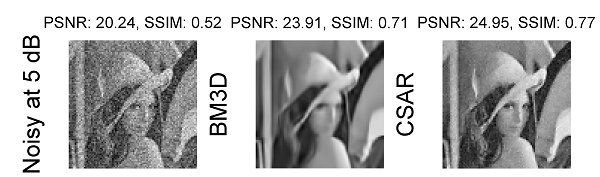}}\quad
	\subfigure{\centering\includegraphics[width=0.49\linewidth]{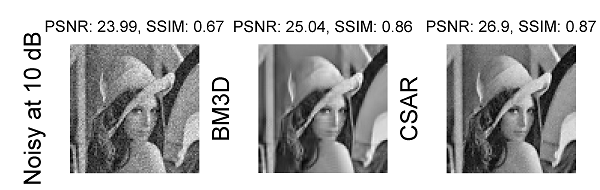}}\\
	\subfigure{\centering\includegraphics[width=0.49\linewidth]{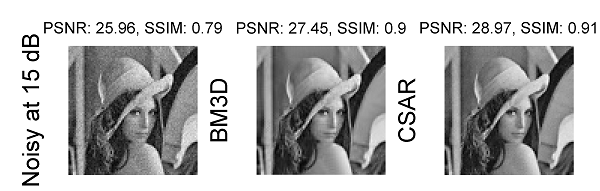}}\quad
	\subfigure{\centering\includegraphics[width=0.49\linewidth]{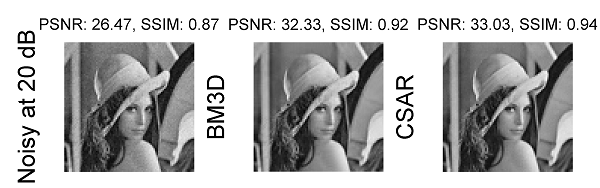}}\\
	\subfigure{\centering\includegraphics[width=0.49\linewidth]{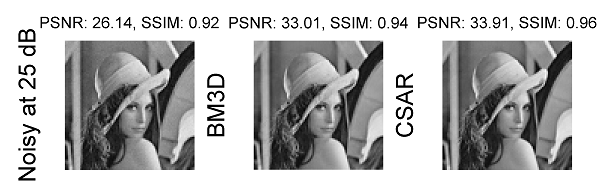}}\\
	\subfigure{\centering\includegraphics[width=0.49\linewidth]{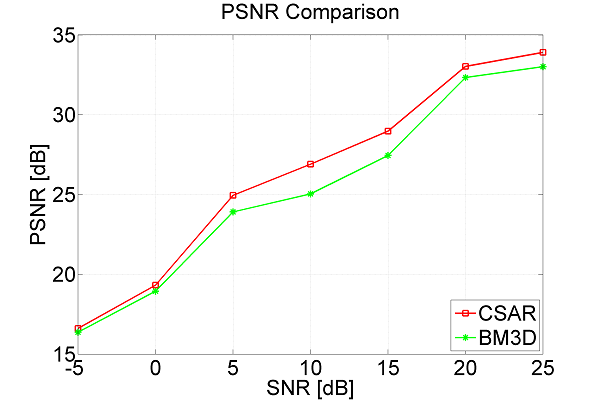}}\quad
	\subfigure{\centering\includegraphics[width=0.49\linewidth]{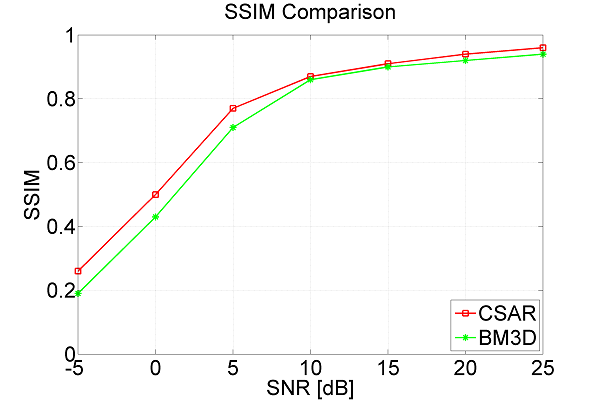}}
	\caption{A comparison of BM3D and CSAR algorithm's based denoising results of Lena standard test image over an extensive SNR range of -5 dB to 25 dB.}
	\label{R_Lena}
\end{figure*}

\begin{figure*}[t!]	
	\subfigure{\centering\includegraphics[width=0.49\linewidth]{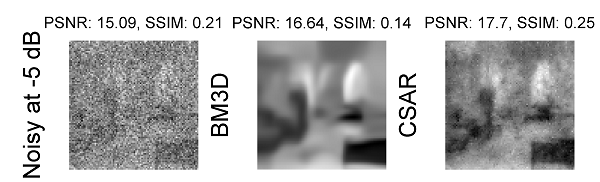}}\quad
	\subfigure{\centering\includegraphics[width=0.49\linewidth]{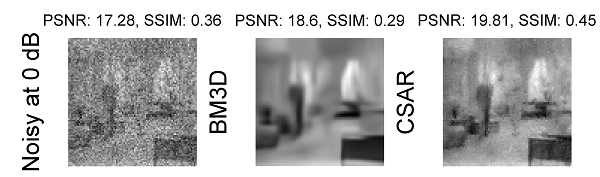}}\\
	\subfigure{\centering\includegraphics[width=0.49\linewidth]{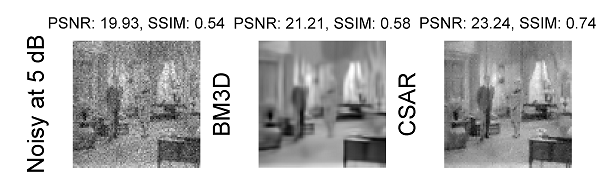}}\quad
	\subfigure{\centering\includegraphics[width=0.49\linewidth]{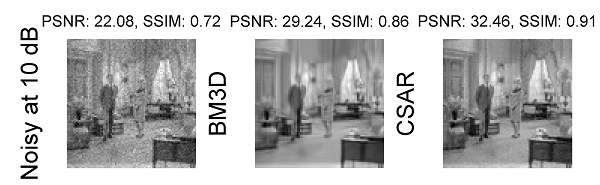}}\\
	\subfigure{\centering\includegraphics[width=0.49\linewidth]{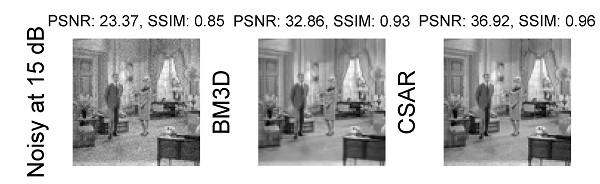}}\quad
	\subfigure{\centering\includegraphics[width=0.49\linewidth]{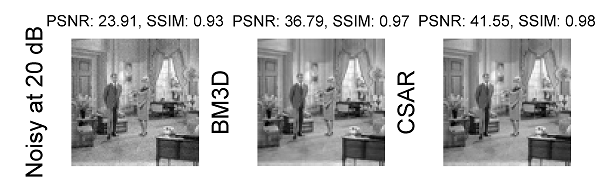}}\\
	\subfigure{\centering\includegraphics[width=0.49\linewidth]{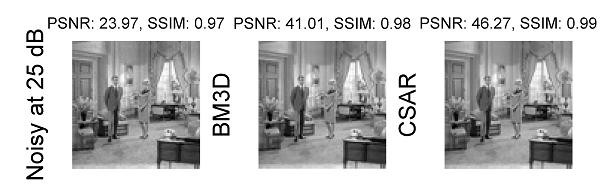}}\\
	\subfigure{\centering\includegraphics[width=0.49\linewidth]{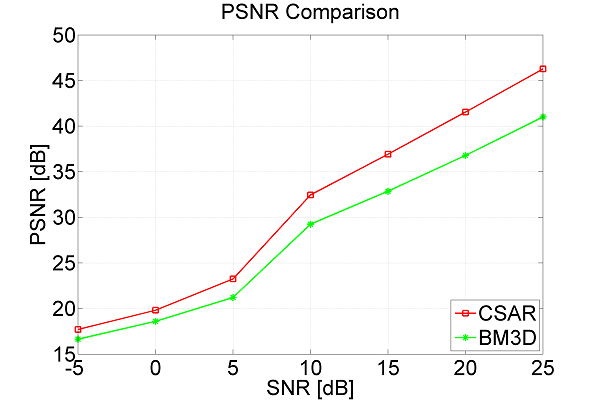}}\quad
	\subfigure{\centering\includegraphics[width=0.49\linewidth]{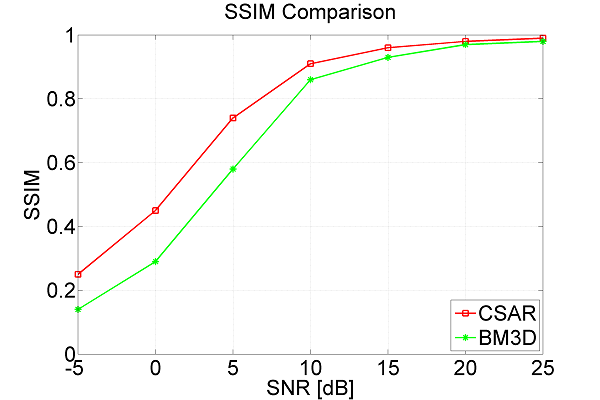}}
	\caption{A comparison of BM3D and CSAR algorithm's based denoising results of Living Room standard test image over an extensive SNR range of -5 dB to 25 dB.}
	\label{R_livingroom}
\end{figure*}

\begin{figure*}[t!]	
	\subfigure{\centering\includegraphics[width=0.49\linewidth]{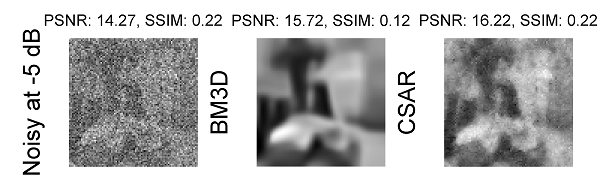}}\quad
	\subfigure{\centering\includegraphics[width=0.49\linewidth]{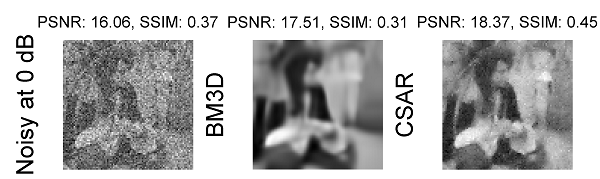}}\\
	\subfigure{\centering\includegraphics[width=0.49\linewidth]{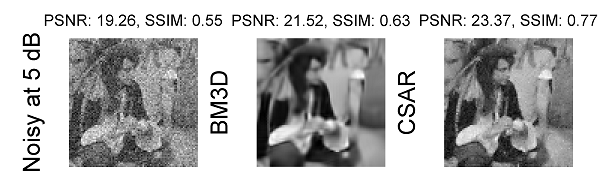}}\quad
	\subfigure{\centering\includegraphics[width=0.49\linewidth]{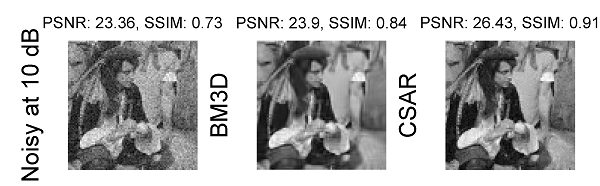}}\\
	\subfigure{\centering\includegraphics[width=0.49\linewidth]{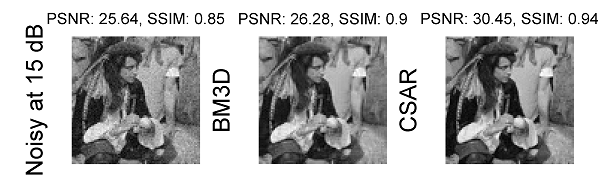}}\quad
	\subfigure{\centering\includegraphics[width=0.49\linewidth]{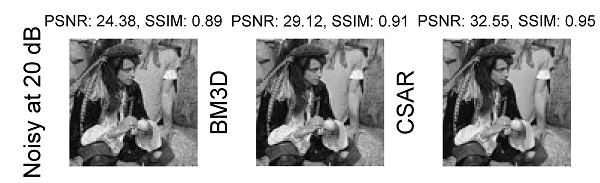}}\\
	\subfigure{\centering\includegraphics[width=0.49\linewidth]{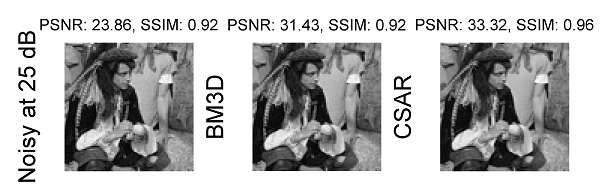}}\\
	\subfigure{\centering\includegraphics[width=0.49\linewidth]{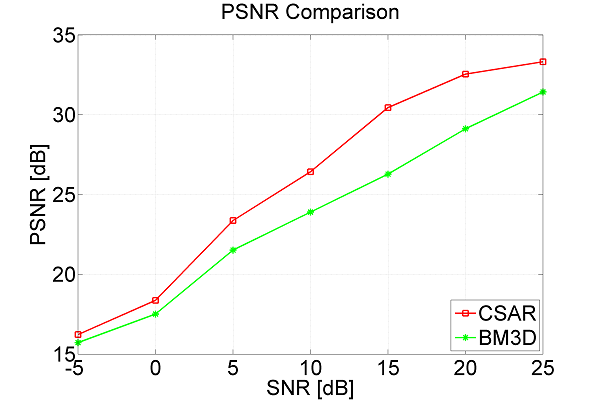}}\quad
	\subfigure{\centering\includegraphics[width=0.49\linewidth]{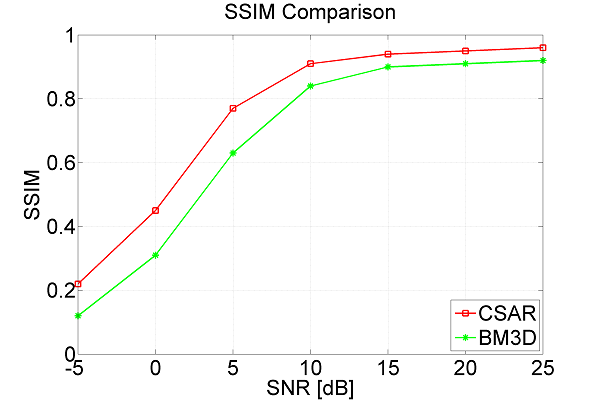}}
	\caption{A comparison of BM3D and CSAR algorithm's based denoising results of Man standard test image over an extensive SNR range of -5 dB to 25 dB.}
	\label{R_man}
\end{figure*}

Further, the results of proposed CSAR algorithm and the competing BM3D algorithm were compared over a large number of standard test images using a wide range of noise levels. For this purpose, we show all the original standard test images in Fig. \ref{all_original} used for the extensive simulations. The noisy and the resulting denoised images using BM3D and proposed CSAR algorithms are shown in Fig. \ref{R_mandrill} to Fig. \ref{R_man}. We show in these figures that our proposed algorithm is capable of both preserving smooth regions of the image as well as the details in the image, which is in fact one of the most challenging tasks while denoising since many denoising algorithms tend to blur out the details. Also as we have compared the results using a wide range of noise levels, this validates that our algorithm is superior to the state-of-the-art algorithm BM3D and is better in terms of both objective and subjective measures.

\clearpage

\clearpage

\clearpage

\clearpage

\clearpage

\clearpage

\clearpage

\clearpage

\clearpage

\clearpage

\clearpage

\section{Conclusion}
\label{Conclusion}
In this paper, we have proposed a novel sparse-recovery-based denoising algorithm. We deploy a patch-based collaborative scheme via enhanced similar patch hunt. The likelihood that a tap is active is computed and refined through collaboration yielding an enhanced sparse estimate, hence improving the isolation of the noise-dominated taps. Results obtained under various experimental setups demonstrate the superiority of the proposed algorithm when benchmarked against selected state-of-the-art algorithms. An interesting future direction is to identify smooth and non-smooth regions to tailor our collaborative framework for further improvements.

\bibliographystyle{ieeetr}
\bibliography{CSAR_Denoising_Detailed}

\end{document}